\documentclass[a4paper,11pt]{article}

\usepackage{fourier}
\usepackage{color}
 \usepackage{graphicx}
\usepackage{url}
\usepackage[affil-it]{authblk}
\usepackage{amsmath}
\usepackage{wrapfig}

\usepackage[T1]{fontenc}
\usepackage{times}

\begin{document}

\title{Decisive Data using Multi-Modality Optical Sensors for Advanced Vehicular Systems}

\author[1] {Muhammad Ali Farooq}
\author[1]{Waseem Shariff}
\author[1]{Mehdi Sefidgar Dilmaghani}
\author[1]{Wang Yao}
\author[2]{Moazam Soomro}
\author[1]{Peter Corcoran}
\affil[1]{School of Engineering, University of Galway, Ireland}
\affil[2]{College of Engineering and Computer Science, University of Central Florida, USA}
\date{}
\maketitle
\thispagestyle{empty}

\begin{abstract}
Optical sensors have played pivotal role in acquiring real world data for critical applications. This data when integrated with advanced machine learning algorithms provides meaningful information thus enhancing human vision.  This paper focuses on various optical technologies for design and development of state-of-the-art out-cabin forward vision systems and in-cabin driver monitoring systems. The focused optical sensors include Long Wave Thermal Imaging (LWIR) cameras, Near Infrared (NIR), Neuromorphic/ event cameras, Visible CMOS cameras and Depth cameras. Further the paper discusses different potential applications which can be employed using the unique strengths of each these optical modalities in real time environment.  
\end{abstract}

\section{Introduction}
Advanced vehicular systems rely on various optical modalities and hardware sensors to enhance driver safety, vehicles efficiency, and performance. Most common key technologies which are directly associated with next generation autonomous vehicles ~\cite{Synopsys} includes.  
\begin{itemize}


\item Camera Vision Systems: Camera-based vision systems capture visual information from inside the vehicle and the vehicle's surroundings. They are used for in cabin applications such as adjusting car internal temperature adjustment according to driver and occupant requirements, lane departure warning, traffic sign recognition, pedestrian detection, and object tracking \cite{1}, \cite{jia2008vision}.  

\item Head-Up Display (HUD) and Augmented Reality (AR): HUD projects important information, such as speed, navigation instructions, and alerts, onto the windshield or a dedicated display unit. It makes easy for the driver to access vital information without taking their eyes off the road, enhancing safety and situational awareness. \cite{charissis2010human}. Similarly AR can be used in vehicle windshields to overlay digital information, such as navigation cues, traffic data, and safety warnings, onto the real-world view. This technology enhances driver assistance and provides a more intuitive driving experience \cite{wang2020augmented}. 

\item Vehicle-to-Everything (V2X) Communication: V2X communication enables vehicles to exchange information with other vehicles, infrastructure, and pedestrians. Optical communication systems, such as visible light communication (VLC), is used to transmit data between vehicles or from traffic signals, enabling real-time cooperative driving, collision avoidance, and traffic management \cite{wang2019survey}. 

\end{itemize}
Keeping this in view in this study we will mainly discuss various camera technologies along with their respective strengths and weakness to elaborate their importance in challenging environmental and atmospheric conditions for advanced driver assistance systems (ADAS) and driver monitoring systems (DMS). Further we have highlighted the importance of machine learning algorithms to extract meaningful and decisive information that can aid drivers in real time driving experience. 

\section{Optical Sensor Technologies and Applications for Vehicular Systems}
In this work we have mainly focused camera vision systems that can effectively employed and integrated with existing vehicular automation systems.

\subsection{Visible CMOS Cameras}
CMOS (complementary metal oxide semiconductor) camera is a digital camera that uses a CMOS image sensor, which is a device that converts light into an electrical signal. Generally, CMOS consists of the following parts: microlenses, color filters, metal lines, photodiodes, and substrates. CMOS cameras have several advantages compared with traditional CCD (charge-coupled device) cameras, including low power consumption, fast readout speeds, high integration, and low cost. These features allow CMOS cameras to be used in a variety of applications as follows.
\begin{itemize}
\item Vehicle Detection and Tracking: CMOS cameras are used to detect and position vehicles~\cite{he2021vehicle} on the road. The captured images or video streams are analyzed by image processing and computer vision algorithms to identify and track vehicles on the road~\cite{do2019visible} for traffic monitoring, vehicle counting, and vehicle speed measurement. 

\item Advanced Driver Assistance Systems (ADAS): CMOS cameras are integral components of ADAS, which is used to warn drivers of any possible hazards on the road including lane departure warnings, forward collision warnings, pedestrian detection, and traffic sign recognition. These cameras capture the road ahead and provide real-time data to ADAS algorithms~\cite{kuo2011vision}, enabling the system to detect and provide assistance to the driver, which is important for improving traffic safety and preventing accidents.

\item Autonomous Vehicle Navigation: CMOS cameras play a crucial role in autonomous vehicle navigation~\cite{liu2008intelligent}. They provide visual input to the autonomous driving system, allowing the vehicle to understand its surroundings, detect and track other vehicles, and make decisions based on the detected objects and their trajectories. 

\item Traffic Signal Control: CMOS cameras can be used in traffic signal control systems to automatically adjust signal timing and dispensing by monitoring vehicle flow~\cite{liu2012traffic} and traffic conditions on the road in real-time to optimize traffic flow and reduce congestion. 
\end{itemize}
Thus, CMOS cameras play a vital role in vehicular systems, providing efficient and accurate image acquisition and processing capabilities, supporting traffic monitoring, safety management, traffic flow analysis, and other applications, and contributing to traffic management and road safety.

\subsection{LWIR Thermal Camera}
LWIR (Long-Wave Infrared) cameras, also known as thermal infrared camera s, are designed to capture thermal radiation emitted by objects. They operate in the long-wave infrared spectrum covering the wavelengths ranging from 8µm to 14µm (8,000 to 14,000nm). Since LWIR cameras works by detecting the thermal infrared radiations also referred to as heat patterns, these cameras do not rely on external lighting conditions which makes them ideal for low lighting scenarios and even zero lightning conditions. The key applications of these camera are as follows.  
\begin{itemize}
\item Night Vision: LWIR cameras can capture images in total darkness or low-light conditions. This feature makes them perfect choice for external roadside environmental monitoring by integrating these cameras with deep learning-based algorithms. We can find various published studies elaborating its extensive usage for external object detection \cite{1}, object tracking, and lane detection \cite{role}. 

\item Vision in Adverse Conditions: LWIR cameras are less affected by factors like smoke, fog, dust, or poor visibility caused by environmental conditions. They can penetrate certain materials and provide visibility in situations where conventional CMOS based visible cameras are unable to produce robust results. This feature allows the drivers to get comprehensive roadside thermal perception even in harsh weather conditions.  

\item Temperature Measurement: LWIR cameras enable non-contact temperature measurement. They can accurately measure the temperature of objects, even in challenging environments or from a distance. This capability is valuable feature for in cabin application such as driver and occupant temperature monitoring thus adjusting internal temperature of the car. Moreover, this feature can be useful for drivers’ drowsiness and fatigue detection thus generating timely alerts to avoid traffic collisions.
\end{itemize}

Figure 1 shows the out-cabin object detection results on various thermal frames acquired from publicly available thermal datasets \cite{9732195}.

\begin{figure}[h]
\centering
\includegraphics[width=0.90\textwidth]{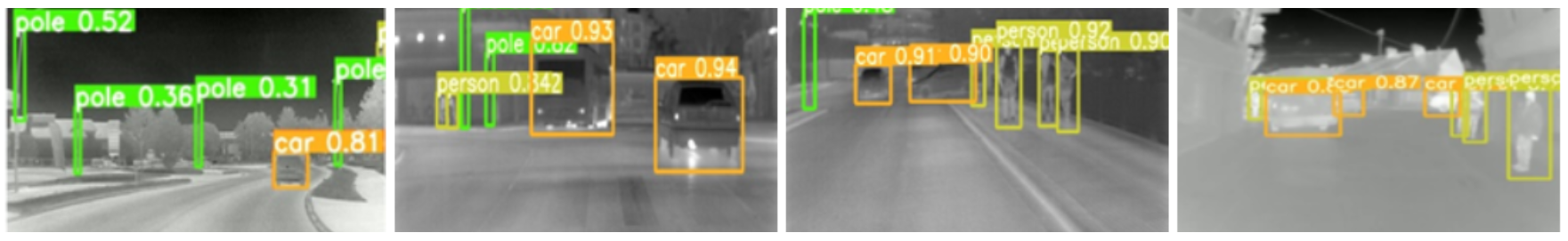}
\caption{Thermal Imaging Frames with Out-cabin Object Detection Results}
\label{fig:1}
\end{figure}
\subsection{NIR Camera}
\begin{wrapfigure}{r}{0.5\textwidth}
  \vspace{-20pt}
  \begin{center}
    \includegraphics[width=0.3\textwidth, height=0.22\textwidth]{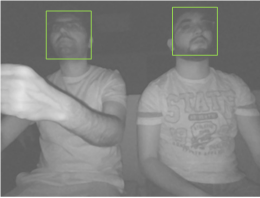}
\end{center}
\vspace{-20pt}
  \caption{NIR output for monitoring the driver and passenger}
  \vspace{-0pt}
\end{wrapfigure}
NIR (Near-Infrared) thermal cameras, also referred to as SWIR (Short-Wave Infrared) thermal cameras, can acquire thermal radiation in the near-infrared spectrum, typically ranging from 0.9 µm to 1.7 µm wavelength spectrum. Likewise, LWIR thermal cameras NIR cameras offer non-contact temperature measurement, and vision through obscuring elements. These features make them optimal choice for different in-cabin and out-cabin vehicular applications such as monitoring driver's attentiveness, detect signs of drowsiness or distraction, and issue alerts and timely warnings. Moreover, NIR cameras can monitor the interior environment of the vehicle as shown in the figure 2. They can detect the presence of occupants, their seating positions, and activity levels. This information can be utilized for personalized climate control, optimizing airbag deployment, or triggering alerts in case of unattended passenger such as children.

\subsection{Neuromorphic Sensors}

Neuromorphic sensors, also known as event cameras, have a unique way of capturing scenes. Unlike traditional RGB cameras that capture multiple frames per second, event cameras have independent pixels that only respond to changes in light, resulting in a stream of events rather than continuous images. Each event includes a timestamp, the pixel's position, and a binary polarity indicating an increase or decrease in light. This type of camera offers advantages such as speed, low memory usage, fewer computations, and privacy preservation, making it ideal for autonomous driving \cite{gracca2023shining, chen2020event}. This research focuses on discussing the in-cabin and out-of-cabin vision applications of event cameras in the automotive industry.

\begin{itemize}
\item In-cabin applications: The event camera's high temporal resolution makes it an excellent choice for applications such as monitoring driver drowsiness in vehicles. Traditional cameras struggle to track subtle facial expressions, blink counting, head pose/gaze estimation, and saccadic movement analysis, which are crucial for assessing drowsiness levels. By utilizing event cameras, vehicles can move closer to achieving full autonomy. Researchers propose new algorithms to monitor drivers, analyze key factors in drowsiness, and address challenges in event camera-based monitoring systems \cite{ryan2021real, dilmaghani2022control}. One approach involves using a LSTM-based system that detects driver distraction by creating 3D tensors from event camera output streams \cite{yang2022event}. Another algorithm focuses on detecting yawning, an important indicator of drowsiness, using frames generated from event streams \cite{kielty2023neuromorphic}.
\item Out-cabin applications: 
Pedestrian detection is crucial for autonomous driving and various approaches have been proposed. In \cite{wan2021event}, a YOLO-v3 based architecture is suggested to detect pedestrians from frames generated by integrating event streams. Another study \cite{ojeda2020device} introduces a hardware-efficient architecture using event streams, consisting of denoising the input stream  and a simple neural network for pedestrian detection. Object detection is also important in next generation automotives, with \cite{wzorek2022traffic} using YOLO-v4 to detect traffic signs, \cite{shariff2022event} presenting a proof of concept for YOLO-based forward perception systems, and \cite{mentasti2022event} proposing two event-based solutions for object detection and tracking in traffic monitoring, one based on geometrical schemes and the other on neural networks.
\end{itemize}

\begin{figure}[htp]
\centering
\includegraphics[width=0.90\textwidth]{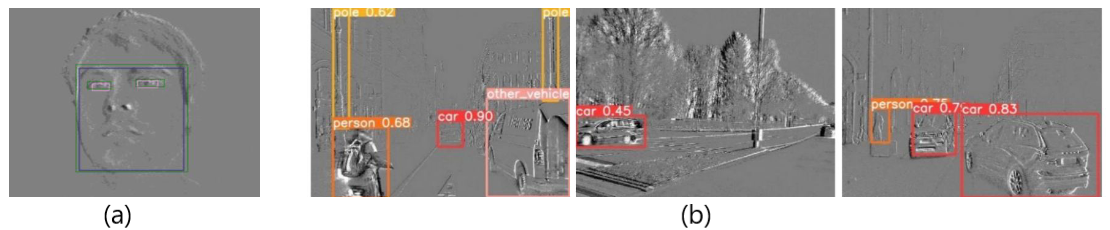}
\caption{Event camera output samples: a) in-cabin driver monitoring \cite{ryan2021real}, b) out-cabin road side object detection \cite{shariff2022event}}
\label{fig:3}
\end{figure}
\subsection{Depth Cameras}
Depth cameras, also known as 3D cameras or depth sensing cameras, are a type of sensor used in automotive applications to capture depth information of the surrounding environment. These cameras operate by emitting infrared (IR) or laser light and measuring the time it takes for the light to bounce back after hitting objects in the scene. By analysing the stereo vision, time-of-flight or structured light patterns, depth cameras can create a three-dimensional representation of the scene \cite{khan2022towards, khan2023robust, kumar2020lidar}. In the context of automotive, depth cameras offer several key features and benefits: 

\begin{figure}[h]
\centering
\includegraphics[width=0.9\textwidth]{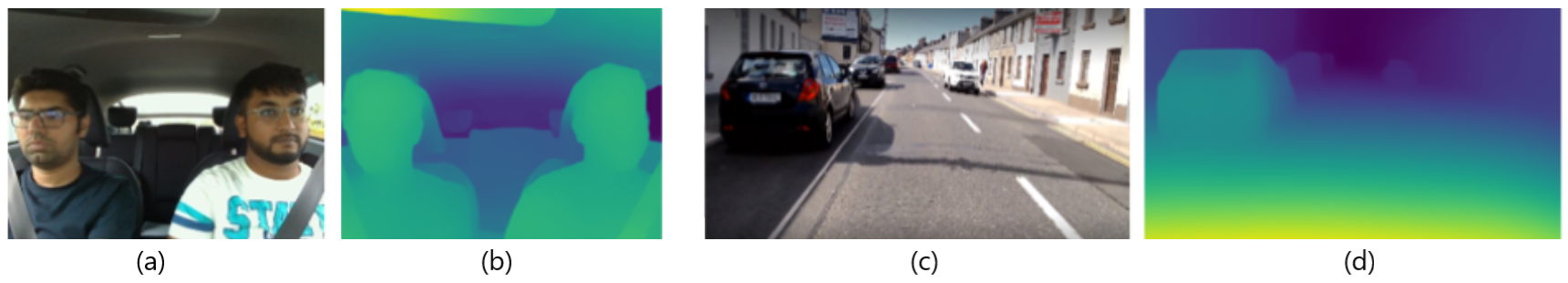}
\caption{Depth Camera Sensor Outputs. a) In-cabin RGB camera b) Corresponding in-cabin depth output c) Out-cabin RGB camera d) Corresponding out-cabin depth output.}
\label{fig:1}
\end{figure}

\begin{itemize}
\item Object Detection with Depth Sensing: Depth cameras provide accurate depth information, allowing for precise detection of moving objects. This enables advanced driver-assistance systems (ADAS) and autonomous vehicles to identify and monitor the position, size, and movement of pedestrians, vehicles, and other obstacles on the road. This information is crucial for making real-time decisions and ensuring safe driving \cite{zhou2022context}. 

\item Distance Measurement: Depth cameras enable accurate measurement of distances between the vehicle and objects in the surrounding environment. This information is essential for adaptive cruise control systems, automatic braking systems, and parking assistance, as it helps determine the appropriate response or manoeuvre required to maintain a safe distance from other vehicles or objects \cite{kumar2020lidar}. 

\item Scene Understanding and Mapping: Depth cameras provide a detailed representation of the scene, allowing vehicles to understand the geometry and structure of the environment. This helps in creating high-definition maps, path planning, and navigation, especially in complex urban environments or during adverse weather conditions where traditional sensors may struggle \cite{mehrzed2023optimization}. 

\item Gesture and Occupant Recognition: Depth cameras can be utilized for driver monitoring and gesture recognition inside the vehicle. By analysing the depth information, these cameras can track the driver's gaze, head position, and hand movements, enabling features such as driver drowsiness detection, distraction monitoring, and intuitive in-car controls through gestures \cite{leu2011novel}. 

\item Enhanced Safety and Collision Avoidance: The depth information captured by these cameras enhances overall safety by providing a more comprehensive understanding of the environment. This enables the development of collision avoidance systems that can detect potential hazards and take proactive measures to prevent accidents \cite{ding2023designs}. 

\item Low-Light Performance: Like LWIR thermal cameras, some depth cameras are designed to work in low-light conditions. They use active illumination, such as infrared light, to capture depth information, making them suitable for driving scenarios with limited visibility or at night \cite{saxena20083}.
\end{itemize}
By combining depth information with other sensor data, such as radar and LiDAR, depth cameras contribute to a comprehensive perception system for autonomous vehicles. They enable accurate and robust scene understanding, enhancing the safety, efficiency, and overall driving experience in automotive applications.

\section{Discussion}
This section will summarizes the potential advantages of different optical modalities along with there weaknesses in automotive scenarios.
\begin{itemize}
\item The visible CMOS sensor provides high-resolution images in good external lighting conditions however the performance of these cameras is severally effected in low-light conditions. The LWIR thermal camera excels in low-light and even in zero-lighting situations by detecting thermal signatures. The NIR camera enhances visibility in low-light, while thermal cameras outperform NIR cameras in complete darkness. The event camera aids in real-time object tracking with low-latency imaging, while the depth camera focuses on depth sensing and not optimized for night vision.

\item For object detection, the visible CMOS sensor enables precise object recognition and classification with its high-resolution color imagery. In challenging lighting conditions, the LWIR thermal camera surpasses visible light cameras by detecting objects based on their thermal emissions. While the NIR camera is capable of object detection, thermal cameras generally outperform NIR cameras due to their ability to sense thermal signatures. The event camera, although primarily designed for real-time tracking, can contribute to overall perception systems by providing valuable visual information. Additionally, the depth camera can offer additional in-depth information, assisting in the identification of objects. 

\item In terms of facial recognition, the visible CMOS sensor may struggle in low-light conditions, while the LWIR thermal camera primarily captures thermal information rather than detailed facial features. The NIR camera is well-suited for facial recognition in low-light scenarios, capturing detailed facial features not easily visible to visible light cameras. The event camera's low-latency imaging capabilities can aid in real-time tracking of facial movements, and the depth camera can provide additional depth information for more accurate analysis of facial characteristics. 

\item For drowsiness and fatigue detection, the visible CMOS sensor has limited capability, while LWIR thermal cameras primarily capture thermal information rather than eye movements or facial features. NIR cameras are well-suited for monitoring driver drowsiness and fatigue due to their ability to analyze facial characteristics and detect eye movements even in low-light conditions. Event cameras have the potential to contribute to drowsiness and fatigue detection with their temporal information feature, but research in this area is limited. Additionally, depth cameras can provide additional depth information for a more comprehensive analysis of the driver's condition. 

\item In the domain of gesture recognition, the visible CMOS sensor can capture gestures but may require proper lighting conditions. LWIR thermal cameras excel in capturing thermal information but are not commonly used for gesture recognition. NIR cameras are well-suited for detailed hand movement capture, even in low-light conditions. Event cameras, with their low-latency imaging capabilities, contribute to real-time tracking of hand movements. Depth cameras, specifically designed for depth sensing, enable precise tracking of hand gestures for gesture recognition. 

\item When it comes to 3D mapping, the visible CMOS sensor has limited capability as it primarily captures color imagery without depth information. LWIR thermal cameras focus on capturing thermal information rather than depth and are not typically used for 3D mapping. Similarly, NIR cameras mainly capture color imagery without depth information and are not primarily used for 3D mapping. Event cameras provide valuable visual information for environment perception, while depth cameras, specifically designed for depth sensing, are ideal for creating accurate 3D maps of the vehicle's surroundings.
\end{itemize}

\section{Conclusion and Future Work}
In summary, the choice of optical sensor depends on the specific requirements for desired application in real-time environment. While visible CMOS sensors offer high-resolution colour imagery, thermal cameras excel in low-light and adverse weather conditions. NIR cameras provide enhanced visibility in low-light situations, event cameras offer real-time tracking capabilities, and depth cameras enable accurate 3D mapping. To make the best choice, it's essential to consider factors such as cost, performance requirements, and the specific needs of the application. By carefully evaluating these factors, one can determine the most appropriate optical sensor that strikes the right balance between functionality and affordability. Ultimately, selecting the right sensor can significantly impact the effectiveness and efficiency of a given system or application. 
As the possible future directions the array of multi modality optical sensors can be deployed in parallel with conventional sensors which includes LIDAR and RADAR for achieving maximum benefits for in-cabin as well as out-cabin applications. 

\bibliographystyle{apalike}

\bibliography{imvip}

\end{document}